\documentclass[runningheads]{llncs}
\usepackage[T1]{fontenc}
\usepackage{graphicx,verbatim, amsmath, amssymb, algorithm, algpseudocode, multirow, booktabs, array, makecell}

\usepackage[colorlinks=true, urlcolor=blue]{hyperref}

\begin{document}
%
\title{Harnessing EHRs for Diffusion-based Anomaly Detection on Chest X-rays}
%

\author{
Harim Kim\inst{1} \and
Yuhan Wang\inst{2} \and
Minkyu Ahn\inst{1} \and
Heeyoul Choi\inst{1} \and
Yuyin Zhou\inst{2} \and
Charmgil Hong\inst{1}
}

\authorrunning{Kim et al.}  

\institute{
Handong Global University, Pohang, South Korea \\
\email{\{hrkim, minkyuahn, hchoi, charmgil\}@handong.ac.kr}
\and
University of California, Santa Cruz, CA, USA \\
\email{\{ywan1332, yzhou284\}@ucsc.edu}
}

\maketitle              
\begin{abstract}
Unsupervised anomaly detection (UAD) in medical imaging is crucial for identifying pathological abnormalities without requiring extensive labeled data. However, existing diffusion-based UAD models rely solely on imaging features, limiting their ability to distinguish between normal anatomical variations and pathological anomalies. To address this, we propose Diff3M, a multi-modal diffusion-based framework that integrates chest X-rays and structured Electronic Health Records (EHRs) for enhanced anomaly detection. Specifically, we introduce a novel image-EHR cross-attention module to incorporate structured clinical context into the image generation process, improving the model’s ability to differentiate normal from abnormal features. Additionally, we develop a static masking strategy to enhance the reconstruction of normal-like images from anomalies. Extensive evaluations on CheXpert and MIMIC-CXR/IV demonstrate that Diff3M achieves state-of-the-art performance, outperforming existing UAD methods in medical imaging.
Our source code is available at \href{https://github.com/nth221/Diff3M}{https://github.com/nth221/Diff3M}.



\end{abstract}

\section{Introduction}
Recent AI advancements have notably improved radiographic image analysis, providing crucial diagnostic support to radiologists \cite{feng2020ultrasound,han2023contrastive,mao2025medsegfactory}.
However, current clinical practices rely on AI tools designed for predefined disease predictions \cite{obuchowicz2025artificial}.
This reliance raises the risk of overlooking clinically significant but previously uncharacterized pathological features.
To overcome this limitation, unsupervised anomaly detection (UAD) offers a promising direction, with the capability to identify undefined pathological features without prior annotation \cite{bao2024bmad}.

Generative model-based anomaly detection has gained interest in recent years \cite{akcay2019ganomaly,schlegl2019f}, distinguishing anomalies by regenerating input data into normal-like representations \cite{schlegl2017unsupervised,schlegl2019f}. 
Among these approaches, diffusion-based models excel in industrial domains, due to their powerful generation capabilities \cite{he2024diffusion,mousakhan2023anomaly}. These models extract semantic features characterizing normal data and use them as a reference to reconstruct the input into a normal-like image.
However, the variability in individual anatomical structures within radiographic images complicates the extraction of features that manifest normal data, making it difficult to enhance detection performance \cite{bercea2024diffusion}. 
To address this challenge, we propose a Diffusion-based Multi-modal Medical Anomaly Detection (Diff3M) framework that leverages both chest X-rays and structured Electronic Health Records (EHRs) for enhanced anomaly detection. While previous studies show that integrating EHR data with medical images enhances clinical prediction \cite{yao2024drfuse,zhou2021radfusion}, its potential in anomaly detection remains largely unexplored. To bridge this gap, we introduce a novel Image-EHR cross-attention (IECA) mechanism that allows EHR data to provide additional semantic context, enabling the model to differentiate between normal anatomical variations and pathological anomalies. This mitigates the impact of anatomical variability in the generation process, resulting in more reliable anomaly detection. Additionally, we propose a novel masking strategy for reconstructing normal-like images. Existing diffusion-based anomaly detection methods primarily employ random masking with predefined textures \cite{iqbal2023unsupervised,zhang2023unsupervised}, which lacks generalizability. In contrast, we introduce a Pixel-level Checkerboard Masking (PCM) module, designed to capture a broader range of anomalies by incorporating static masking patterns that generalize beyond specific textures. Throughout the reverse diffusion process, Diff3M reconstructs masked regions into normal-like images by leveraging both the noised input and EHR-guided semantic embeddings, enhancing its ability to detect and characterize anomalies effectively.

We conduct comprehensive experiments using two major chest radiography datasets, CheXpert \cite{irvin2019chexpert} and MIMIC-CXR/IV \cite{johnson2019mimic,johnson2024mimic}, both offering rich collections of chest X-rays paired with clinical information.
Our evaluation, combining quantitative metrics and qualitative analysis, shows that the proposed method outperforms existing UAD approaches for medical images.
The contributions of this paper can be summarized as follows:
\begin{enumerate}
\item The first diffusion-based medical anomaly detection framework that effectively integrates EHR data as conditioning.
\item A novel image-EHR cross-attention module to generate EHR embeddings that incorporate structured clinical information into the detection pipeline.
\item A new static masking strategy to enhance the reconstruction of normal-like images from anomalies.
\item State-of-the-art performance in medical UAD, surpassing existing methods.
\end{enumerate}

\section{Background}
\label{sec:bg}
\noindent\textbf{Diffusion-based UAD}
Diffusion-based models reconstruct anomalies into nor-\\mal-like images. 
AnoDDPM \cite{wyatt2022anoddpm} improves normal-like image generation using Simplex noise \cite{perlin2002improving} but underperforms compared to standard denoising autoencoders \cite{fan2024survey}.
mDDPM \cite{iqbal2023unsupervised} applies random masking but targets specific anomalies rather than undefined abnormalities. 
DDAD \cite{mousakhan2023anomaly} and the more recent DiAD  \cite{he2024diffusion} condition the reverse process on extracted image features, which may mix normal and anomalous traits, leading to generalization issues.
\\

\noindent\textbf{Feature-based UAD}
Feature-based UAD methods detect anomalies by analyzing features extracted from pre-trained foundation models \cite{bao2024bmad}.
PaDiM, CFA and PatchCore \cite{defard2021padim,lee2022cfa,roth2022towards} use memory bank to compare features from normal training and testing data, while
RD4AD \cite{deng2022anomaly} and MambaAD \cite{he2024mambaad} employ knowledge distillation to detect anomalies by measuring discrepancies between teacher and student networks.
However, the general-purpose foundation models may struggle to extract meaningful medical features.
\\

\noindent\textbf{Diffusion Models}
The Denoising Diffusion Probabilistic Model (DDPM) \cite{ho2020denoising} consists of two main processes: a forward process $q(\cdot)$ and a reverse process $p_\theta(\cdot)$.
The forward process gradually adds Gaussian noise to input data $\textbf{x}$ over $T$ steps, mapping it to a standard normal distribution.
Each step $t$ ($t =1,..., T$) introduces noise $\epsilon^{(t)}\sim N(0,1)$, scaled by a variance parameter $\beta_t \in (0, 1)$ like $q(\mathbf{x}_t | \mathbf{x}_{t-1}) = \mathcal{N}(\mathbf{x}_t; \sqrt{1 - \beta_t} \mathbf{x}_{t-1}, \beta_t \mathbf{I}).$
The reverse process is a learnable procedure that reconstructs data from $\textbf{x}_T$ to generate sample $\textbf{x}_0$ resembling the training data.
It is trained by minimizing the KL divergence of $q(\mathbf{x}_{t-1} | \mathbf{x}_t, \mathbf{x}_0)$ and $p_{\theta}(\mathbf{x}_{t-1} | \mathbf{x}_t)$.
It can be interpreted as approximating each step of the reverse process as the inverse operation of the corresponding forward process step. 
Consequently, the model learns to predict the noise $\epsilon^{(t)}$ using $    \mathcal{L}:=\|\epsilon_{\theta}^{(t)}(\mathbf{x}_t) - \epsilon^{(t)}\|^2_2.$

The Denoising Diffusion Implicit Model (DDIM) \cite{song2020denoising} enhances DDPM efficiency by reformulating the forward process with a reversed ordinary differential equation (ODE), making it deterministic and non-Markovian \cite{wolleb2022diffusion}:
\begin{equation}
\textbf{x}_{t+1} = \textbf{x}_t + \sqrt{\bar{\alpha}_{t+1}} \left[ \left( \sqrt{\frac{1}{\bar{\alpha}_t}} - \sqrt{\frac{1}{\bar{\alpha}_{t+1}}} \right) \textbf{x}_t + \left( \sqrt{\frac{1}{\bar{\alpha}_{t+1}}} - 1 - \sqrt{\frac{1}{\bar{\alpha}_t}} + 1 \right) \epsilon_\theta^{(t)} \right].
\end{equation}
where $\bar{\alpha}_t:=\prod_{s=1}^{t} (1-\beta_s)$.
Since the processes are no longer Markovian, DDIM enables sample generation with fewer steps than $T$.
The method proposed in this paper is built upon DDIM.

\section{Proposed Framework}
We present \textbf{Diff3M}, a diffusion-based UAD framework trained exclusively on normal data. It comprises two key steps: (1) a \emph{Noise Prediction} (NP) Network for estimating noise $\epsilon_\theta^{(t)}$ in the reverse diffusion process and (2) a \emph{Masked Pixel Generation} (MPG) Network for reconstructing masked inputs into normal-like images $\tilde{\mathbf{x}}_t$. To effectively integrate EHRs for enhanced anomaly detection, we propose two key designs. 
First, we introduce a novel \emph{Image-EHR Cross Attention} (IECA) module, which extracts EHR embeddings from paired X-ray images \( \mathbf{x} \) and EHR data \( \mathbf{r} \), conditioning the MPG Network and the NP Network (Sec. \ref{ehram}).
This enables a more semantically informed diffusion process. Additionally, following the original DDPM framework \cite{ho2020denoising}, we incorporate timestamp embeddings to capture temporal information. Second, we propose a \emph{Pixel-level Checkerboard Masking} (PCM) strategy to enhance image reconstruction by generating masked inputs from \( \mathbf{x}_t \), which the MPG Network then reconstructs into normal-like images (Sec. \ref{pcm}).

\begin{figure}[t]
\centering
\includegraphics[width=0.9\textwidth]{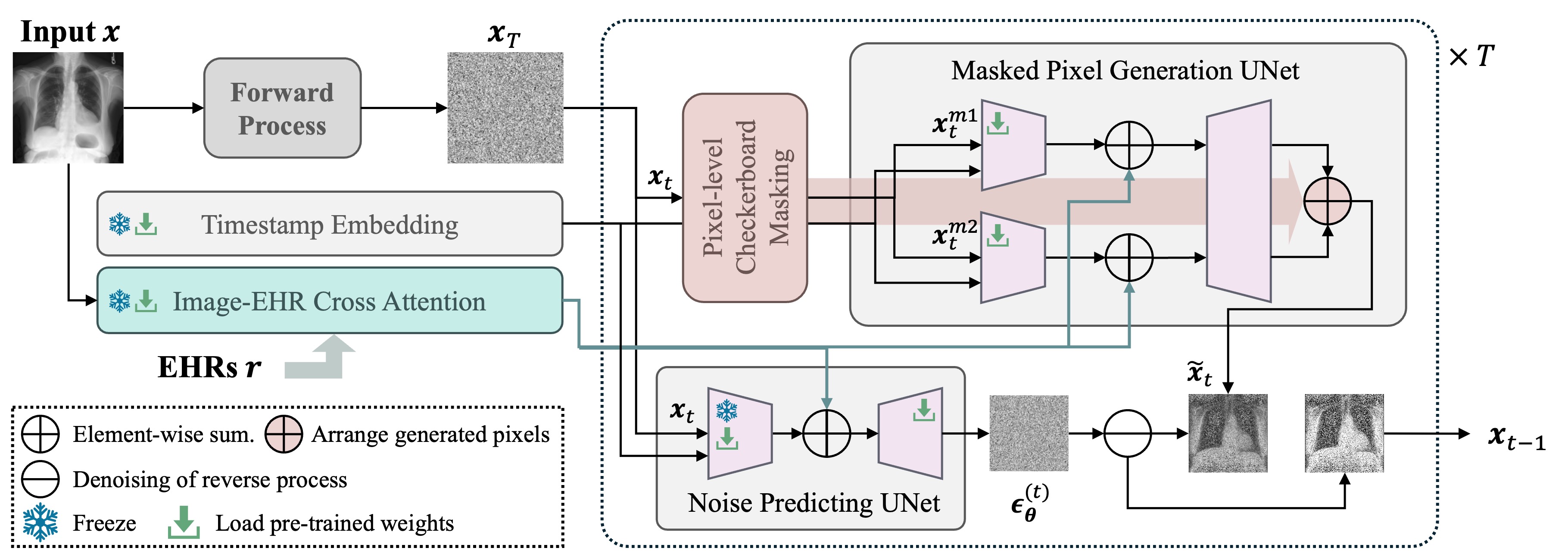}
\vspace{-0.5em}
\caption{Training process of Diff3M. Image-EHR Cross Attention (IECA) incorporates EHR embeddings as conditional inputs for the noise prediction (NP) network at each reverse process step. 
A novel Pixel-level Checkerboard Masking (PCM) strategy is employed to improve generating  masked inputs into normal-like images from the Masked Pixel Generation (MPG) Network.
} 
\vspace{-1em}\label{fig:training_framework}
\end{figure}

\subsection{Image-EHR Cross Attention (IECA)} \label{ehram}
\begin{figure}[t]
\centering
\includegraphics[width=0.85\textwidth]{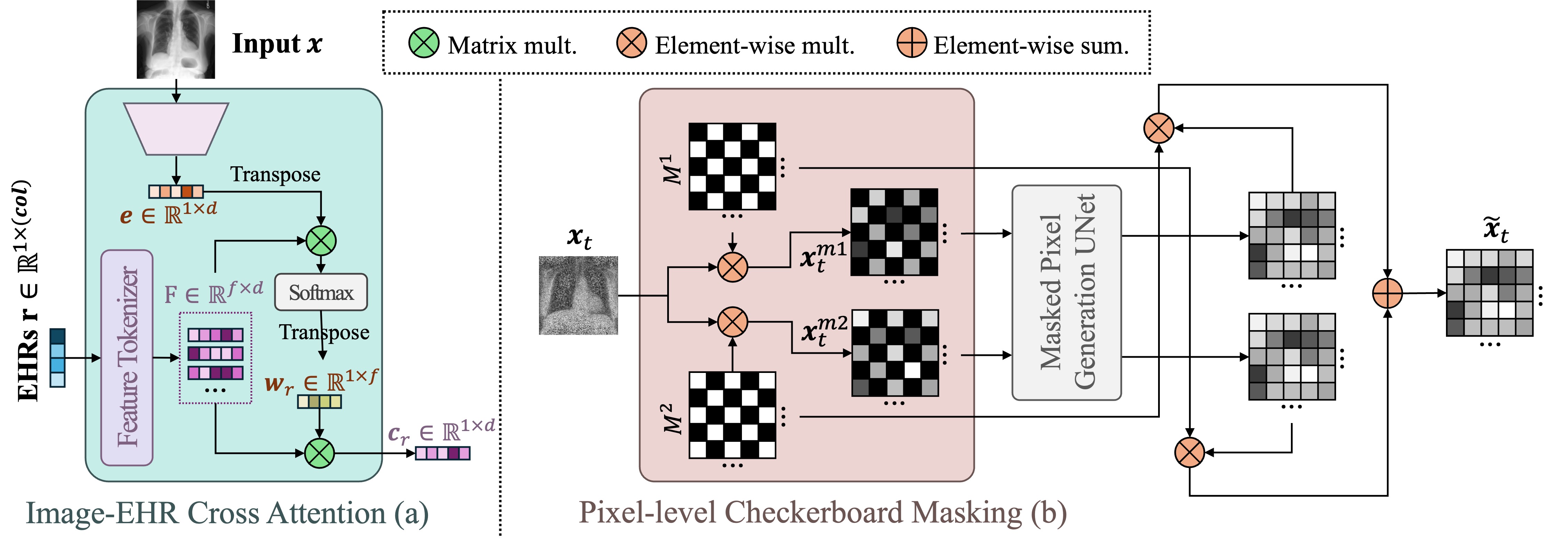}
\vspace{-1em}
\caption{Details of the proposed modules. (a) The Image-EHR Cross Attention module  incorporate structured clinical context into the image generation process. (b) The Pixel-level Checkerboard Masking module facilitates the regeneration of anomalies into normal-like images using two types of static masks.} \label{fig:modules}
\end{figure}

Existing diffusion-based models \cite{he2024diffusion,mousakhan2023anomaly} extract normal semantic features solely from input images, potentially incorporating anatomical anomalies. To mitigate this, we propose an Image-EHR Cross Attention (IECA) module to derive an embedding that represents the X-ray image through EHR data, reducing the influence of anomalous features.
As shown in Fig. \ref{fig:modules}-(a), we first use a feature tokenizer \cite{gorishniy2021revisiting} to generate EHR embeddings, producing $f$ tokens based on tabular  features.
The generated tokens $F \in \mathbb{R}^{f \times d}$ are then used to compute token-wise weights $\textbf{w}_r$ based on their similarity to the input image embedding $\textbf{e}$.
The final conditional embedding \( \textbf{c}_r \) is then obtained as a weighted sum of \( F \):
\begin{equation}
\begin{aligned}
    &F = \text{FeatureTokenizer}(\textbf{r}), \quad \textbf{r} \in \mathbb{R}^{1 \times (\text{col})}, \quad F \in \mathbb{R}^{f \times d},\\
    &\textbf{c}_r= \textbf{w}_rF, \quad  \textbf{w}_r = \sigma_{softmax}((F \textbf{e}^T)/\sqrt{d}),\quad \textbf{e} = \text{Encoder}(\textbf{x}), \quad \textbf{e} \in \mathbb{R}^{1 \times d}\\
\end{aligned}
\end{equation}
Since the generated conditional embedding $\textbf{c}_r$ is based on the similarity with $\textbf{e}$, it implicitly encodes the EHRs information to align with the semantic features of the input image.
$\textbf{c}_r$ is then used as a conditional input for MPG and NP to provide normal semantic features for the input image.

\subsection{Pixel-level Checkerboard Masking (PCM)} \label{pcm}
Existing methods apply random masking with specific textures  to simulate predefined target anomalies, training models to reconstruct the patterns into normal-like anatomical shapes.
However, such approaches often struggle to detect real pathological abnormalities that deviate from predefined patterns.
Instead, we employ static masking to deliberately obscure image regions, forcing the model to rely on auxiliary information—specifically, the EHR conditional embedding from IECA—to reconstruct normal-like images.
This approach enables the detection of a broader range of anomalies beyond those seen in training data.


As illustrated in Fig. \ref{fig:modules}-(b), PCM consists of two complementary checkerboard masks ($M^1,M^2$) of input image size.
A scale parameter $s$ adjusts masking intensity by time step $t$ to prevent later losses affecting earlier steps.
At time step $t$, applying $M^1$ and $M^2$ to noised input $\textbf{x}_t$ yields $\textbf{x}_t^{m1}$ and $\textbf{x}_t^{m2}$.
To obtain the final prediction, we apply the opposite mask to each reconstructed image.
The formulation is as follows:
\begin{equation}
\begin{aligned}
    M^1_{i,j} &= (i + j) \bmod 2, \quad M^2_{i,j} = 1 - (i + j) \bmod 2,\\
    &M^{1,2} \gets (M^{1, 2} \times (1-s) + s), \quad s=t/T,\\
    \mathbf{\tilde{x}}_t = \text{MPG}(\textbf{x}&_t^{m1}) \odot M^2 + \text{MPG}(\textbf{x}_t^{m2}) \odot M^1,\quad \textbf{x}_t^{m1}, \textbf{x}_t^{m2}= \textbf{x}_t \odot M^{1,2}
\end{aligned}
\end{equation}
The regenerated $\mathbf{\tilde{x}}_t$ is used for loss function during training and for generating the previous step image $\textbf{x}_{t-1}$ during detection.

\subsubsection{Training and Detection}
To train the noise prediction capability of the NP Network and the masked image reconstruction capability of the MPG Network, we compute the mean squared error (MSE) between output of each network ($\epsilon_\theta^{(t)}$, $\mathbf{\tilde{x}}_t$) and its corresponding target ($\epsilon^{(t)}$, $\textbf{x}_t$) as follows:
\begin{equation}
\mathcal{L}_\textit{Diff3M} = \mathbb{E}_{t, \mathbf{x}, \mathbf{c}_r, \epsilon} \left[ \lambda \|\epsilon_{\theta}^{(t)} - \epsilon^{(t)}\|_2^2 + (1-\lambda) \| \mathbf{\tilde{x}_t} - \mathbf{x}_t\|_2^2 \right]
\end{equation}
where $\lambda \in (0, 1)$ is a weighting factor.
For more efficient training, the base DDPM and IECA module can be pre-trained on X-ray images and EHR data (Fig. \ref{fig:training_framework}).


Algorithm \ref{alg:anomaly_detection} outlines the Diff3M detection process.
Using the DDIM sampling strategy, the forward process introduces noise into the image up to $T'$ ($T'<T$) steps.
During the reverse process, the model generates $\textbf{x}_{t-1}$ using produced noise $\epsilon_\theta^{(t)}$ and image $\mathbf{\tilde{x}}_t$
following \cite{wolleb2022diffusion}.
The final anomaly score measures the difference between the reconstructed image $\textbf{x}_0$ and the original input $\textbf{x}$.
Samples with high anomaly scores are flagged as potential anomalies.

\begin{algorithm}
    \caption{Anomaly detection using Diff3M}
    \label{alg:anomaly_detection}
    \begin{algorithmic}[1]
        \State \textbf{Input:} input image $\mathbf{x}$, EHRs $\mathbf{r}$, timestamp $t$, noise level $T'$
        \State \textbf{Output:} image-level anomaly score $a$
        
        \For{all $t$ from $0$ to $T'-1$}
            \State $\textbf{x}_{t+1} \gets \textbf{x}_t + \sqrt{\bar{\alpha}_{t+1}} \left[
            \left( \sqrt{\frac{1}{\bar{\alpha}_t}} - \sqrt{\frac{1}{\bar{\alpha}_{t+1}}} \right) \textbf{x}_t
            + \left( \sqrt{\frac{1}{\bar{\alpha}_{t+1}}} - 1 - \sqrt{\frac{1}{\bar{\alpha}_t}} - 1 \right) 
            \epsilon_{\theta}^{(t)} \right]$
        \EndFor
        \State $\mathbf{c}_r \gets IECA(\mathbf{r}, \mathbf{x})$
        \For{all $t$ from $T'$ to $1$}
            \State $t_{emb} \gets TimestampEmbedding(t)$
            \State $\mathbf{\tilde{x}}_t \gets MPG(\mathbf{x}_t^{m1},\mathbf{x}_t^{m2}, \mathbf{c}_r, t_{emb}), \quad(\mathbf{x}_t^{m1},\mathbf{x}_t^{m2})\gets PCM(\mathbf{x}_t)$
            \State $\textbf{x}_{t-1} \gets \sqrt{\bar{\alpha}_{t-1}} \left(
            \frac{\mathbf{\tilde{x}}_t - \sqrt{1 - \bar{\alpha}_t} {\epsilon}_\theta^{(t)}}{\sqrt{\bar{\alpha}_t}} \right) + \sqrt{1 - \bar{\alpha}_{t-1}} {\epsilon}_\theta^{(t)},\quad\epsilon_\theta^{(t)} \gets NP(\textbf{x}_t, \textbf{c}_r, t_{emb})$
        \EndFor
        \State $a \gets diff(\textbf{x}, \textbf{x}_0), \quad diff(\textbf{x}, \textbf{x}_0) \in  \{\| \textbf{x} - \textbf{x}_0 \|_2^2 , \,\, \underset{i,j}{\max} | \textbf{x}^{i, j} - {\textbf{x}}_0^{i,j} | \}$
        \State \textbf{return} $a$
    \end{algorithmic}
\end{algorithm}

\section{Experiments and Results}
\noindent\textbf{Datasets and Evaluation Metrics}
CheXpert \cite{irvin2019chexpert} and MIMIC-CXR \cite{johnson2020mimic} both contain X-ray images, the same types of demographic information, and 12 disease classes.
We define normal samples as cases with \textit{No Finding}, while all other cases are considered anomalies.
For demographic features, we use sex, age, and AP/PA view. In MIMIC-CXR, we further incorporate BMI, blood pressure, height, and weight from MIMIC-IV \cite{johnson2020mimic} as additional EHR features, selecting records within three months of the X-ray imaging date.
CheXpert has 16,969 normal samples for training, and (26/176) (normal/anomaly) samples for testing.
MIMIC has 29,310 normal samples for training, and (487/1646) (normal/anomaly) samples for testing. 
We evaluate image-level anomaly detection using AUROC and AUPRC to compare performance across various detection thresholds.
\\

\noindent\textbf{Implementation Details}
Diff3M is trained using 8 Nvidia RTX A5000 GPUs.
All images are resized to 256$\times$256.
IECA, MPG, and NP adopt the UNet-based architecture from \cite{wolleb2022diffusion}.
We use Adam optimizer with a maximum time step $T$ of 1,000, a batch size of 3, a weighting factor $\lambda$ of 0.5 and a learning rate of $10^{-4}$.
For Chexpert, the model is trained from scratch for 60,000 iterations.
For MIMIC, the model is pre-trained on CheXpert, and fine-tuned with EHRs for 70,000 iterations.
During inference, DDIM sampling is applied with 600/400 steps for CheXpert/MIMIC.

 
\subsection{Performance Comparison}
We compare Diff3M with state-of-the-art (SOTA) feature-based models, e.g., PatchCore \cite{roth2022towards}, RD4AD \cite{deng2022anomaly}, and MambaAD \cite{he2024mambaad}, along with the diffusion-based model DiAD \cite{he2024diffusion}.
For qualitative evaluation, we compare the anomaly maps, computed as the absolute pixel-level difference between the input and output.
\\
\begin{table}[t]
\caption{Performance comparison on CheXpert and MIMIC datasets. $demo$: demographics information for IECA. $ehr$: additional EHRs from MIMIC-IV for IECA.}\label{tab:comparison}
\centering
\setlength{\tabcolsep}{3pt}
\renewcommand{\arraystretch}{0.6}
\fontsize{8pt}{9pt}\selectfont
\begin{tabular}{l|c|ccc|ccc}
\toprule
\multirow{2}{*}{Dataset} & \multirow{2}{*}{Metric} & \multicolumn{3}{c|}{Feature-based} & \multicolumn{3}{c}{Diffusion-based} \\
\cmidrule{3-8}
 &  & \makecell{PatchCore} & RD4AD & \makecell{MambaAD} & DiAD & \makecell{Ours$_{demo}$} & \makecell{Ours$_{demo,ehr}$} \\
\midrule
\multirow{2}{*}{CheXpert} & AUROC & 0.584 & 0.529 & 0.594 & \textbf{0.664} & \textbf{0.664} & - \\
 & AUPRC & 0.888 & 0.719 & 0.889 & 0.917 & \textbf{0.931} & - \\
\midrule
\multirow{2}{*}{MIMIC} & AUROC & 0.573 & 0.555 & 0.570& 0.597 & 0.610 & \textbf{0.617} \\
 & AUPRC & 0.796 & 0.800 & 0.779 & 0.803 & 0.818 & \textbf{0.821} \\
\bottomrule
\end{tabular}
\end{table}

\begin{figure}[t]
\centering
\includegraphics[width=0.9\textwidth]{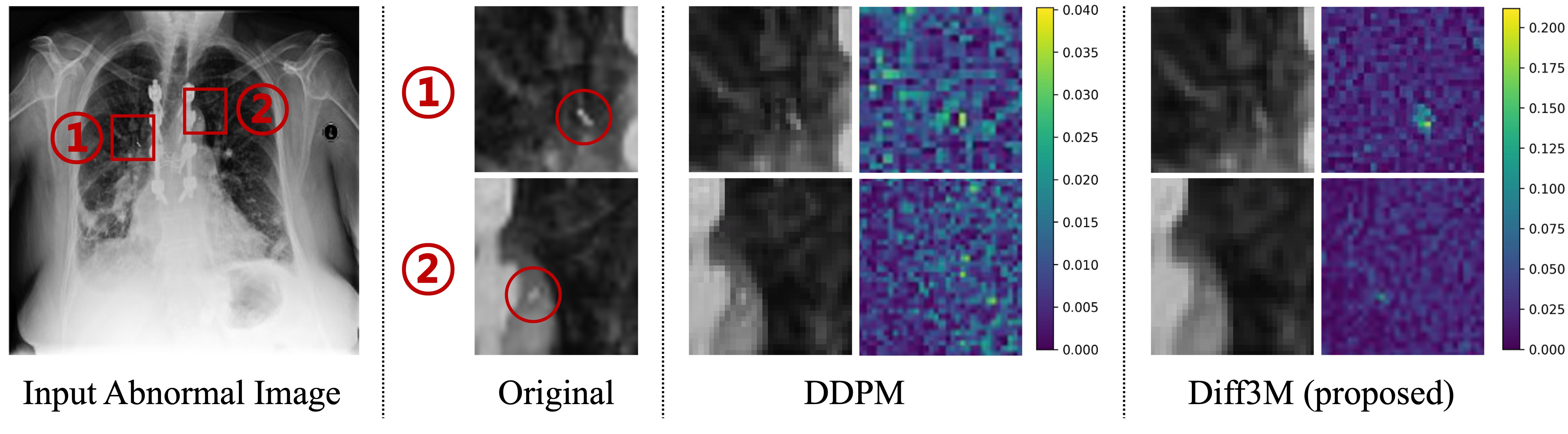}
\vspace{-1em}
\caption{Qualitative comparison. Red circles indicate anomalous artifacts.} \label{fig:heatmap}
\vspace{-1em}
\end{figure}

\noindent\textbf{Quantitative and Qualitative Results}
As shown in Table \ref{tab:comparison}, Diff3M achieves SOTA performance in image-level anomaly detection for both CheXpert and MIMIC. For MIMIC, Diff3M improves AUROC by approximately 2.0–5.5\% over previous approaches. 
This highlights the importance of effectively leveraging EHR data to enhance anomaly detection.
For CheXpert, the performance of Diff3M using only demographics-based EHRs is comparable to DiAD. 
This suggests that demographic information alone may not fully capture anatomical variations.
Overall, diffusion-based approaches outperform feature-based methods, likely because foundation models struggle to extract effective features in the medical domain, as discussed in Sec.~\ref{sec:bg}.

Fig. \ref{fig:heatmap} compares the anomaly regeneration capability of DDPM and Diff3M. We examine an abnormal image containing statistically anomalous artifacts. 
Unlike DDPM, Diff3M successfully transforms anomalous artifacts into normal-like structures for each region, assigning significantly higher anomaly scores compared to surrounding areas.
These results demonstrate that Diff3M effectively 
generates anomaly maps, which can serve as a valuable medical assistive tool.

\subsection{Ablation Study}
Table \ref{tab:ablation} shows that adding PCM and IECA sequentially improves performance on both datasets.
This demonstrates that each module contributes to anomaly detection and that their combination further enhances performance.
We also compare the performance of MSE and pixel-level maximum absolute difference ($max_{abs}$) as anomaly scoring methods. The results show that MSE is more effective for MIMIC, whereas $max_{abs}$ performs better for CheXpert. To understand this discrepancy, we analyze anomaly detection performance for individual disease classes. In MIMIC, switching to $max_{abs}$ leads to a 12.3–19.9\% performance drop for \textit{Fracture} and \textit{Pleural Other} cases. Conversely, in CheXpert, the number of positive samples for these classes is extremely low (0–1), which explains why using $max_{abs}$ does not negatively impact performance.
These findings suggest that different scoring functions may be more suitable for different diseases.
Therefore, when using the anomaly detection models as a medical assistive tool, it may be necessary to review anomalies based on different scoring functions.

\begin{table}[t]
\caption{Ablation study on the modules of Diff3M (MSE/$max_{abs}$).}\label{tab:ablation}
\centering
\setlength{\tabcolsep}{3pt}
\renewcommand{\arraystretch}{0.6}
\fontsize{8pt}{9pt}\selectfont
\begin{tabular}{l|c|cccc}
\toprule
Dataset & Metric & DDPM & DDPM+PCM & Ours$_{demo}$ & Ours$_{demo,ehr}$\\
\midrule
\multirow{2}{*}{CheXpert} & AUROC & 0.500/0.500 & 0.553/0.640 & 0.547/\textbf{0.664} & - \\
 & AUPRC & 0.838/0.867 & 0.902/0.922 & 0.899/\textbf{0.931} & - \\
\midrule
\multirow{2}{*}{MIMIC} & AUROC & 0.562/0.551 & 0.592/0.562 & 0.610/0.567 & \textbf{0.617}/0.568 \\
 & AUPRC & 0.800/0.791 & 0.807/0.793 & 0.818/0.799 & \textbf{0.821}/0.798\\
\bottomrule
\end{tabular}
\end{table}

\begin{table}[t]
\caption{Analysis of attention weights for each input feature (mean$\pm$std). $entire$: analysis across entire samples. $top_{10\%}$: analysis for the top 10\% of feature values.}\label{tab:attention}
\centering
\setlength{\tabcolsep}{5pt}
\fontsize{8pt}{9pt}\selectfont
\begin{tabular}{c|l|c|c|c|c|c}
\toprule
\multicolumn{2}{c|}{input feature $f$}& BMI & BP$_{max}$ & BP$_{min}$ & Height & Weight \\
\midrule
\multirow{2}{*}{\makecell{mean \\ $\textbf{w}_r^{(f)}$}} & $entire$ & \textbf{0.262}$\pm$0.117 & 0.183$\pm$0.043 & 0.116$\pm$0.037 & 0.210$\pm$0.040 & 0.230$\pm$0.038 \\
\cmidrule{2-7}
 & $top_{10\%}$ & \textbf{0.475}$\pm$0.033 & 0.124$\pm$0.022 & 0.081$\pm$0.024 & 0.145$\pm$0.021 & 0.174$\pm$0.013 \\
\bottomrule
\end{tabular}
\end{table}
\vspace{-0.5em}

\paragraph{\textbf{Effects of EHRs}} \label{eff_pr}
Table \ref{tab:attention} presents the mean of attention weights ($\textbf{w}_r$) assigned to each IECA input feature $f$ in the MIMIC test dataset.
Among the EHR features, BMI has the highest average weight.
In particular, for the top 10\% of samples, BMI accounts for 47.5\%, indicating a significantly dominant contribution.
This suggests that BMI is the most representative feature for capturing the anatomical characteristics of input images.
Unlike height and weight, which independently provide limited information about body shape, BMI (calculated as weight/height$^2$) offers a more meaningful approximation of body shape.
These findings demonstrate that IECA effectively combines EHR features, extracting conditional embeddings that reflect anatomical information.
In the case of demographic features, most of the weights converged to zero.
These results suggest that utilizing richer EHR features is crucial for robust anomaly detection.

\section{Conclusion}
This paper proposes Diff3M, a novel framework that leverages EHR data for enhancing medical anomaly detection. By integrating EHR data through Image-EHR Cross Attention (IECA) and enforcing structured feature learning with Pixel-level Checkerboard Masking (PCM), Diff3M consistently outperforms existing diffusion-based and feature-based methods on CheXpert and MIMIC datas-ets. Future work will explore
incorporating richer clinical data to further enhance its diagnostic support capabilities.
\bibliographystyle{splncs04}
\bibliography{mybibliography}
%




\end{document}